\documentclass[conference]{IEEEtran}
\IEEEoverridecommandlockouts
\usepackage{cite}
\usepackage{amsmath,amssymb,amsfonts}
\usepackage{algorithmic}
\usepackage{textcomp}
\usepackage{times}
\usepackage[none]{hyphenat}
\usepackage{url}
\usepackage{makecell}
\usepackage{latexsym}
\usepackage{array}
\usepackage{lipsum}
\usepackage{booktabs}
\usepackage{subfigure}
\usepackage{multirow}
\usepackage{xcolor}
\usepackage{float}
\usepackage{hyperref}
\hypersetup{colorlinks = true,
            linkcolor = .,
            urlcolor  = purple,
            citecolor = green,
            anchorcolor = blue}

\usepackage{graphicx}
\graphicspath{{./}}
\usepackage{caption}
\makeatletter
 \let\old@ps@headings\ps@headings
 \let\old@ps@IEEEtitlepagestyle\ps@IEEEtitlepagestyle
 \def\confheader#1{%
 \def\ps@headings{%
 \old@ps@headings%
 \def\@oddhead{\strut\hfill#1\hfill\strut}%
 \def\@evenhead{\strut\hfill#1\hfill\strut}%
 }%
 \def\ps@IEEEtitlepagestyle{%
 \old@ps@IEEEtitlepagestyle%
 \def\@oddhead{\strut\hfill#1\hfill\strut}%
 \def\@evenhead{\strut\hfill#1\hfill\strut}%
 }%
 \ps@headings%
 }
 \makeatother

 \usepackage[pscoord]{eso-pic}

\begin{document}
\title{LR-Net: A Block-based Convolutional Neural Network for Low-Resolution Image Classification}

% \author{
% \IEEEauthorblockN{1\textsuperscript{st} Ashkan Ganj}
% \IEEEauthorblockA{\textit{Department of Electrical  and Computer Science} \\
% \textit{University of Mohaghegh Ardabili} \\
% Ardebil, Ardebil, Iran \\
% AshkanGanj@gmail.com}
% \and
% \IEEEauthorblockN{2\textsuperscript{nd} Mohsen Ebadpour}
% \IEEEauthorblockA{\textit{Department of Electrical  and Computer Science} \\
% \textit{Amirkabir University of Technology} \\
% Tehran, Tehran, Iran \\
% mohsenebadpour@outlook.com}
% \and
% \IEEEauthorblockN{\indent 3\textsuperscript{rd} Mahdi Darvish}
% \IEEEauthorblockA{\textit{Department of Electrical  and Computer Science} \\
% \textit{University of Mohaghegh Ardabili} \\
% Tehran, Tehran, Iran \\
% mahdydarvish@gmail.com}
% \and
% \IEEEauthorblockN{4\textsuperscript{th} Hamid Bahador}
% \IEEEauthorblockA{\textit{Department of Electrical  and Computer Science} \\
% \textit{University of Mohaghegh Ardabili} \\
% Ardebil, Ardebil, Iran \\
% hamid.bahador@uma.ac.ir}

% }

\author{\IEEEauthorblockN{$\text{Ashkan Ganj} ^ 1$, $\text{Mohsen Ebadpour} ^ 2$, $\text{Mahdi Darvish} ^ 1$, $\text{Hamid Bahador}^ {1, *}$ \\[0.2cm]}
\IEEEauthorblockA{\textsuperscript{1}Department of Electrical  and Computer Engineering, University of Mohaghegh Ardabili, Ardabil, $\text{Iran}$\\ \textsuperscript{2}Department of Computer Engineering, $\text{Amirkabir University of Technology, Tehran, Iran}$} \\
 
\textit{* Corresponding author: hamid.bahador@uma.ac.ir}
}

\maketitle

\begin{abstract}
The success of CNN-based architecture on image classification in learning and extracting features made them so popular these days, but the task of image classification becomes more challenging when we apply state of art models to classify noisy and low-quality images. It is still difficult for models to extract meaningful features from this type of image due to its low resolution and lack of meaningful global features. Moreover, high-resolution images need more layers to train which means they take more time and computational power to train. Our method also addresses the problem of vanishing gradients as the layers become deeper in the deep neural networks that we mentioned earlier. In order to address all these issues, we developed a novel image classification architecture, composed of blocks that are designed to learn both low-level and global features from blurred and noisy low-resolution images. Our design of the blocks was heavily influenced by Residual Connections and Inception modules in order to increase performance and reduce parameter sizes. We also assess our work using the MNIST family datasets, with a particular emphasis on the Oracle-MNIST dataset, which is the most difficult to classify due to its low-quality and noisy images. We have performed in-depth tests that demonstrate the presented architecture is faster and more accurate than existing cutting-edge convolutional neural networks. Furthermore, due to the unique properties of our model, it can produce a better result with fewer parameters.
The source code of the project is available at this \href{https://github.com/AshkanGanj/Block-Based-ImageClassification-Architecture}{Github repository.}  
\end{abstract}


\section{Introduction}
\label{sec:introduction}
In the last yeras Deep convolutional neural networks changed the era of computer vision. They can be used in most of the computer vision tasks such as object detection \cite{8951147, 8237584}, image classification \cite{7298594,7780459}, or instance image segmentation \cite{8237584}. It is better to use a convolutional neural network rather than a feed-forward network since it allows parameter sharing and dimensionality reduction. In CNN, the feature of parameter sharing leads to reduction of the number of parameters, therefore, the computations are also reduced. Based on CNNs, the principal idea is that pixels and their surroundings have semantic meaning within each image, while elements of interest can appear anywhere within the image. Similarly to MLPs, CNN's have layers, but they're not fully connected: they've got filters, which are sets of cube-shaped weights that are applied to the whole image.

Since AlexNet's \cite{10.5555/2999134.2999257} remarkable success in the ILSVRC-2012 image classification competition, which combined GPU and CNN, further research has focused on enhancing the CNN architecture and integrating it with new concepts to get higher performance.VGG\cite{simonyan2014very}, GoogleLeNet\cite{szegedy2015going}, and ResNet\cite{he2016deep} are three popular attempts to improve performance by using CNNs.To improve performance, the Visual Geometry Group (VGG)\cite{simonyan2014very} is looking into deeper network performance by extracting as many features as possible from high-resolution photos, or the Inception module network (GoogleNet)\cite{szegedy2015going} is attempting to perform multiple operations (pooling, convolution) in parallel with different filter sizes (3x3, 5x5, etc.) to reduce the possibility of making trade-offs. For example, in CNN models, the choice of pooling stages and deciding about the kernel size are two critical trade-offs that have a great impact on the outcome. ResNet\cite{he2016deep} generates residual learning blocks by employing identity mapping shortcut connections. As a result, with hundreds or even thousands of layers, the neural network model can overcome the gradient vanishing problem. Furthermore, DenseNet\cite{huang2017densely} and others found that reorganizing connections between layers of a network improves learning and representation.

All of the models mentioned above have one thing in common: they work better with high-resolution images. However, the issue is that high-resolution images are not always possible due to image age or bandwidth and computation limitations. After testing some of the most advanced models, such as Inception-v3\cite{szegedy2016rethinking}, with low-resolution images, such as Oracle-MNIST\cite{wang2022oracle}, which contain a lot of natural noise as well, we observed a performance degradation in models compared with their results with high-resolution images. Because DNNs and CNNs are built on the same architecture, this problem appears to be common\cite{dodge2016understanding}. Overall, it has been recognized that poor image quality has a significant impact on the performance of deep neural networks in computer vision tasks, as noted here\cite{Koziarski2018ImpactOL}. Many factors influence the quality of an image in the real world. Most of the time, we cannot get a pure and high-resolution image, and most images have some level of noise and degradation, which will diminish the final result. So in this paper, we try to directly address this problem with a new architecture by utilizing the idea of inception\cite{szegedy2015going} to get as many features as possible from images by using different kernels and combining them with some residual connections\cite{he2016deep} to solve problems like vanishing gradients and the problem of dimensionality in deep neural networks.

This paper commences with introducing some related works in the following section. Next, it explains the approach proposed and presented the experimental setup, and talk about the result of training in sections \textcolor{red}{\ref{sec:method}} and \textcolor{red}{\ref{sec:exper}}, respectively, and finally, we have a conclusion in section \textcolor{red}{\ref{sec:con}}.
\section{Related Work}
The use of Convolutional Neural Networks (CNNs) has been a key method of recognizing images, such as classifying them \cite{krizhevsky2012imagenet, szegedy2015going}, recognizing actions\cite{gkioxari2015contextual}, and locate objects \cite{bell2016inside}. As deep learning models require a lot of training instances to be able to converge, Pre-trained models have been implemented to process small- and medium-sized datasets. There seems to be a noticeable improvement in accuracy with the method mentioned above. Nevertheless, because of the pretrained weights to large datasets (e.g., ImageNet\cite{deng2009imagenet}), it is more time consuming and computationally intensive than ever. 

According to Han et al.\cite{xiao2017/online} and Wang et al.\cite{wang2022oracle}, in order to challenge deep learning models, they created benchmark datasets that are acceptable and share the same characteristics of MNIST, namely that the datasets are small in size and encoded in an easy way to use. Images from both Fashion and Oracle are converted into a format compatible with the MNIST dataset. Thus, they can use MNIST's original dataset instantly regardless of the machine learning system. Also, noted datasets contain more complex information than simple digits extracted from the MNIST.

 Several innovative models for classifying $2\times*28$ images were presented in the literature \cite{hirata2020ensemble}. In order to characterize the images in the fashion-MNIST dataset using convolutional neural networks, the authors prepared three types of neural networks. The model exhibits amazing results on the benchmark dataset. An extensive correlation was established between various CNN structures (for example, VGG16) on various datasets (for example, Image Net) using the leNet-5 network designed for fashion-MNIST. As one example, a custom CNN type with stacked convolution layers of VGG 16 achieved an accuracy rate of 93.07 percent on the Fashion MNIST in its published study \cite{greeshma2019hyperparameter}. Various models of CNNs were introduced to determine which of them is most suitable for characterization and identification in terms of their accuracy. The deep learning architectures that were applied were LeNet-5, AlexNet, VGG-16, and ResNet. 

In most cases, the models perform exceptionally well on specific data, but they do not generalize well to similar datasets. As an example, \cite{lei2020shallow} proposed a shallow convolutional neural network using batch normalization techniques in order to accelerate training convergence and improve accuracy. The noted network consists of only four layers with small convolution kernels, resulting in a low time and space complexity. Even though the model achieved top accuracy on the digits MNIST dataset\cite{deng2012mnist}, it was not able to perform sufficiently on both CIFAR\cite{krizhevsky2012imagenet} and Fashion MNIST\cite{xiao2017/online}.

It is the intention of most of the recently developed Deep convolutional neural networks (DCNNs) that they utilize Inception and Residual connections as the basis to implement bigger deep networks. In order to make the model more accurate and detailed, the parameters of the architecture are increased substantially as the size and depth of the model increases. The complex nature of this training increases the complexity of the model, which, in turn, increases the amount of resources required to run it. Recurrence is a difficult property to incorporate in popular Inception architectures, but it is crucial to improving training and testing accuracy by requiring fewer computational resources. Some researchers have attempted to implement more complex DCNN architectures such as GoogleNet \cite{szegedy2015going}, or residual networks with 1001 layers \cite{he2016identity} that are capable of high recognition accuracy when applied to different benchmark datasets.

As we intend to tackle the problem of handling the low resolution image and its classification, we consider the CNN's first layers as feature extractors, after that, images are classified by taking advantage of these features. In order to maximize efficiency, a custom CNN block is designed. As compared to mainstream DCNN architectures, this model not only guarantees a higher recognition accuracy while requiring fewer computation parameters but also contributes to the overall training process of the deep learning approach as a whole.
\section{Method}

\label{sec:method}
\subsection{problem overview}
In this section, we are going to talk about the reason that we decided to introduce a new architecture for image classification. After doing some tests with some famous state-of-the-art models on the Oracle-MNIST\cite{wang2022oracle}, we find out a noticeable decrease in the performance of the models. With some further experiments, we realized that the relation between classiﬁcation accuracy and the resolution of images is close to being linear, in all the cases\cite{article}, and this problem seems to be common in most image classification algorithms. Even images with small degradation or noise, which is not noticeable by humans can cause performance problems for models. According to our findings, we believe that the primary cause of this problem is the architecture of the models, and in order to solve the issue, we believe that new algorithms and architectures need to be developed. In this paper, we introduced a new architecture for addressing this problem.
\begin{figure}
    \centering
    \includegraphics[height = 0.53\textwidth]{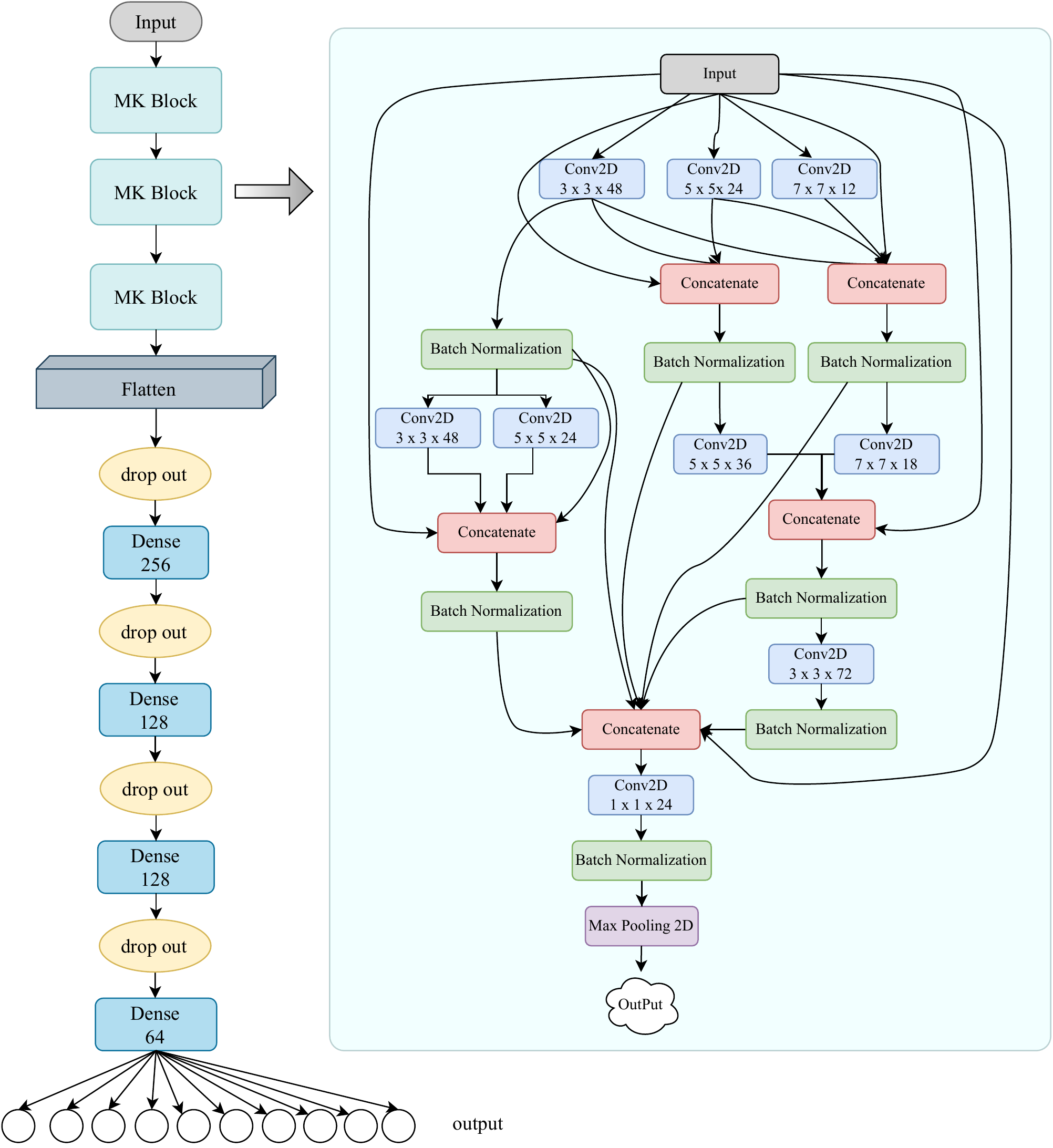}
    \caption{An illustration of the overall architecture of the model is shown in this figure, along with details on each of the blocks. Our block architecture is influenced by both Inception and Residual Net concepts. The block can be divided into two different sides, from which we can extract specific features. As part of our approach, three kernels ($3 \times 3$, $5 \times 5$, $7 \times 7$) were used for feature extraction and a $1 \times 1$ kernel was used for feature combining at the last step.}
    \label{fig:Architecture}
\end{figure}
\subsection{Overall Architecture}
As you can see from the figure. \textcolor{red}{\ref{fig:Architecture}}, our architecture consists of 3 Multi-kernel blocks that stack on top of each other, including the steps that our models take to learn the features and details. After that, we add some Fully connected layers and the last layer with a sigmoid activation function for classifying the outputs. The proposed architecture uses the concept of inception and residual connections in MK blocks to provide robust performance for classifying images. As we know the inception modules were introduced as a way to reduce computational expenses in CNN, and we knew that the simple models couldn't solve this problem so we tried to get the inception module idea and use it with a combination of residual connections to improve accuracy and reduce computation overhead. Since our neural network deals with a wide variety of images with varying salient parts or features, so using an inception base idea is essential for this architecture. 

As our input images are low-resolution images, the useful details usually exist in fewer pixels, and for each window or kernel, the information that can be extracted is rarely found, so the filter size in our models is descending, which means that for larger sizes, we will have smaller kernel sizes. Additionally, the padding of each convolutional layer is the same since the results must be concatenated.
\begin{figure}[!hb]
    \centering
    \includegraphics[height = 0.4\textwidth]{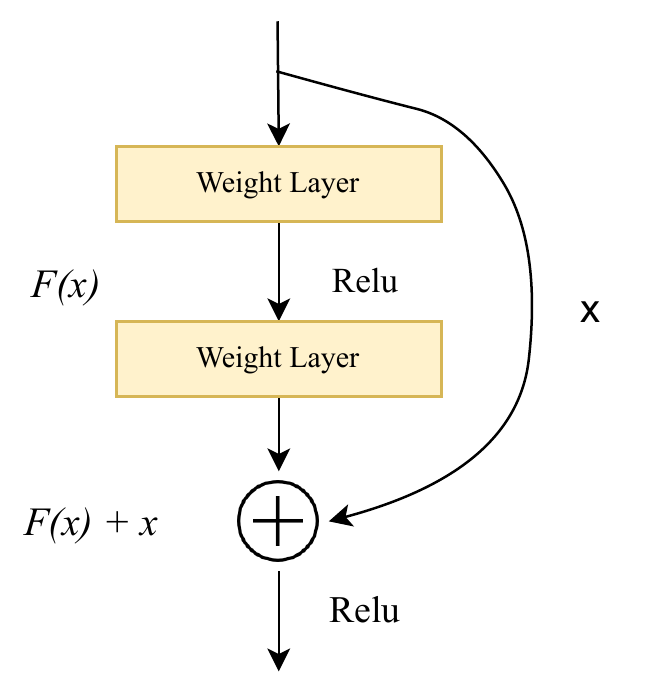}
    \caption{The building block of the deep residual network.}
    \label{fig:res-conn}
\end{figure}

The Multi-Kernel block (MK-block) contains several residual connections, as shown in \textcolor{red}{\ref{fig:Architecture}}. Rather than learning unreferenced functions, these links learn residual functions by referencing the layer inputs. 
The stacked nonlinear layers are let to fit another mapping of $F(x):=H(x)\times x$ that corresponds to the desired underlying mapping $H(x)$. $F(x)+x$ is formed from the initial mapping (see \textcolor{red}{\ref{fig:res-conn}}). The residual mapping is generally easier to tweak than the original one. In theory, fitting an identity mapping by a stack of nonlinear functions requires less effort than pushing the residual to zero if an identity mapping is optimal.

\subsection{Multi-Kernel block}

\begin{figure}[ht]
    \centering
    \includegraphics[height = 0.3\textwidth]{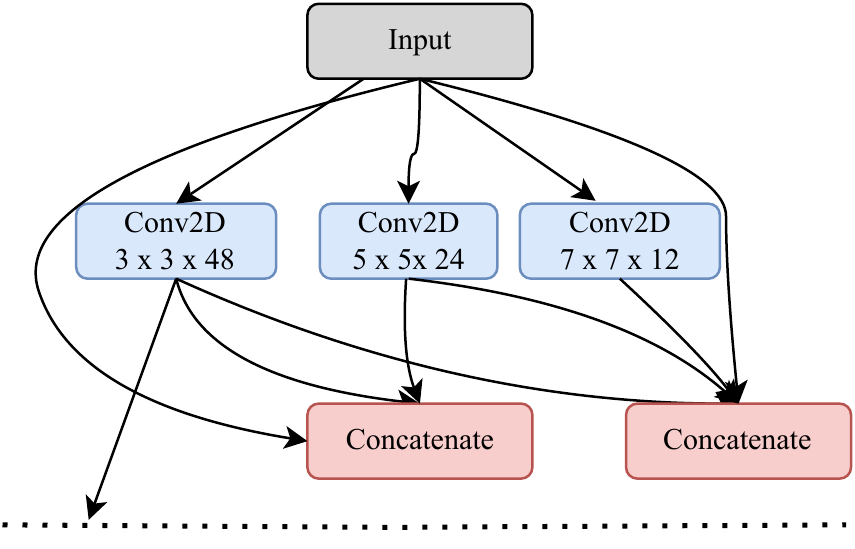}
    \caption{MK-block's first layer shows details of kernel connections as well as how information is shared between them.}
    \label{fig:block-layer1}
\end{figure}

As we mentioned earlier, our model was constructed by stacking 3 Multi-Kernel blocks(Mk-kernel) on top of each other. Our architecture relies heavily on these identical blocks. At the first layer(figure. \textcolor{red}{\ref{fig:block-layer1}}) of our block, we have 3 different kernels ($3 \times 3$, $5 \times 5$, $7 \times 7$). The connection of these 3 kernels is critical.Because, we want to share the information in a way that our model doesn't experience a big jump in kernel and feature. In other words, we connect the result of kernels that have meaningful relationships between them. For example, we have a connection between kernels $3 \times 3$ and $5 \times 5$ because the information that they exchange is useful. However, we do not have a connection between $3 \times 3$ and $7 \times 7$, because the information that they will share is unusable, and the second reason to have this type of connection is that we want to process the low-resolution information. This is because as we know, the big kernels extract more global information, while smaller kernels extract detail and local information. If you look at figure 1, you will see that we have a connection between $3 \times 3$, $5 \times 5$, and $7 \times 7$. The reason we made this connection is that the $5 \times 5$ kernel can create a balance between the $3 \times 3$ and $7 \times 7$ images in a way that the model will better understand which parts of the image will have a local view and which parts will have a global view.

We divide our block into two parts in the second layer. On the right side, we connect the nodes with kernel $5 \times 5$ and kernel $7 \times 7$ together, and at this step, the model attempts to conclude the information in big kernels without considering details, and it will use this information at the last layer when we want to aggregate all the features. Our next layer is a $3 \times 3$ Conv layer that gathers the last details from small sizes, or we can say that we will do local features extraction or low-level features extraction based on global features. The model will try to extract as much detail and low-level features as it can in the continuation of conv2D with kernel $3 \times 3$, since in the previous layer we extracted some low-level features, and so we repeat this operation again. The reason we need this step is that we need good performance on noisy and blurry images. At the end of the process, the extracted features should be applied with any weight from the previous steps.

The last point to note is the way we chose the number of filters. In our images, we have two types of features some of them are local and some of them are global features. When we choose small filters, the model will extract the small and detailed features, which will perform poorly with low-quality and noisy images, while when we choose larger kernels, a much more complex neural network will be needed to extract all useful images. As a result, we decided to implement the inception concept, which uses different image sizes across layers. Because our goal is to classify the low-resolution images. So, we decided to have more $3 \times 3$ kernels than $7 \times 7$ and $5 \times 5$ kernels.
\section{Experiments}
\label{sec:exper}

\subsection{Datasets}
\label{Datasets}
\begin{table}[ht]
\captionsetup{justification=centering}
\caption{Information regarding the size and number of instances of datasets used for evaluation and training in this paper.}
\centering
\setlength{\tabcolsep}{4pt}
\begin{tabular}{p{0.1\textwidth}*{4}{l{c}}}
\toprule
\multicolumn{1}{l}{\multirow{2}{*}{ \makecell{Dataset name}}} & \multicolumn{2}{c}{Training data} & \multicolumn{2}{c}{Validation data} & \multicolumn{2}{c}{Test data} \\
\cmidrule(r{3.8pt}){2-3} \cmidrule(l){4-5} \cmidrule(l){6-7}
& Instances &   Size  &  Instances &   Size & Instances &   Size  \\
\midrule
\makecell{Digit \\MNIST}                & --    & 64.73 & --    & 88.41 & --     & 94.13\\\addlinespace
\makecell{Fashion \\MNIST\cite{xiao2017/online}}            &  71.1    & 63.32   & 36.4 & 88.70 & 20.7 & 94.93\\\addlinespace
\makecell{Oracle \\MNIST\cite{wang2022oracle}} & \textbf{49.4} & \textbf{68.55}  & \textbf{22.3} & \textbf{91.73} & \textbf{14.1} & \textbf{96.28} \\

\bottomrule
\end{tabular}
\label{table:dataset}
\end{table}  

We use MNIST-family datasets to evaluate our work results and compare them with other state-of-art models to show the superiority of our model. The MNIST family datasets have both the characteristics we want to study which are noise and low-quality image. The table \textcolor{red}{\ref{table:dataset}} shows a summary all the information provided in the next parts.
\begin{figure*}[ht]
    \centering
    \includegraphics[width = \textwidth]{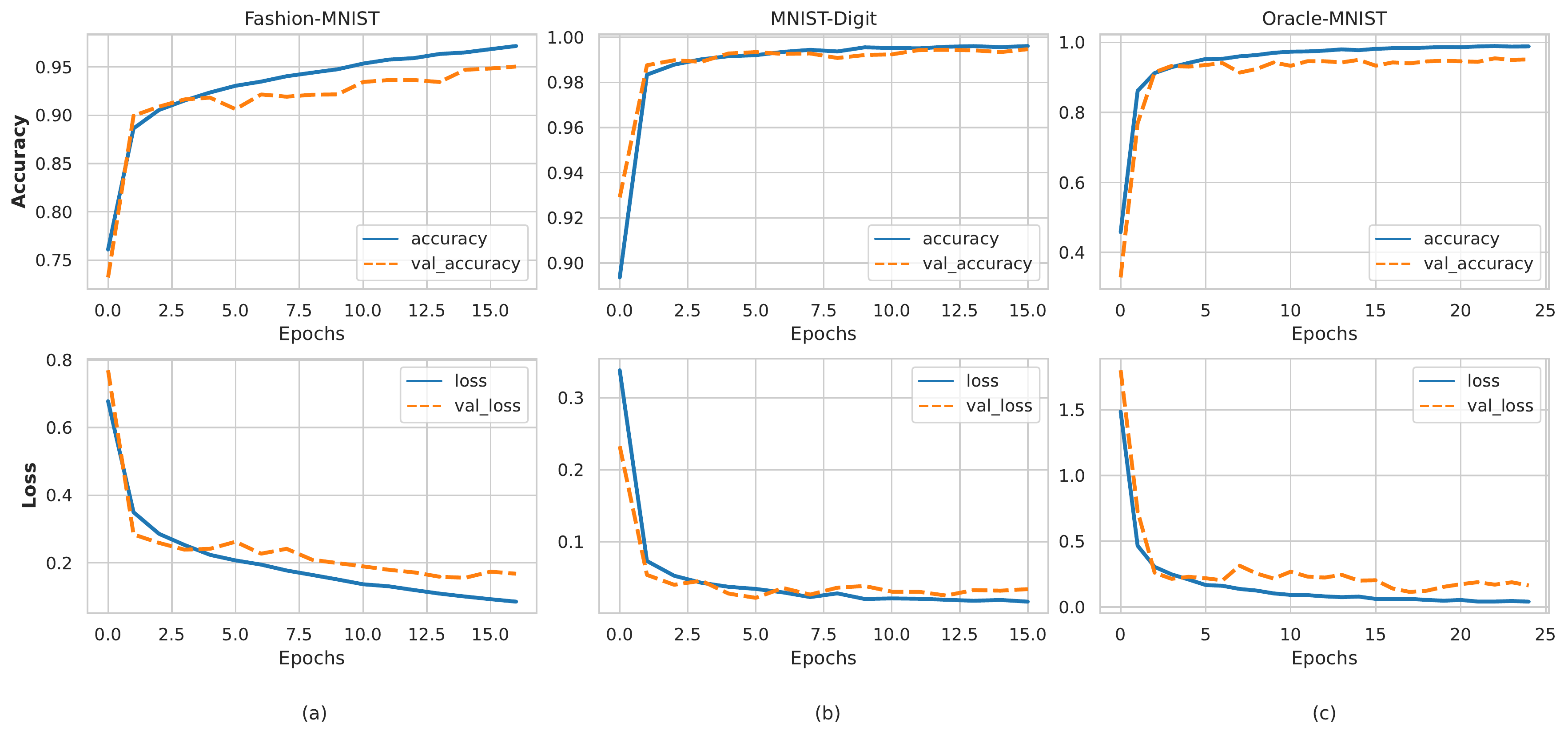}
    \caption{Training and validation loss over epochs generated by training the custom CNN model on (a) MNIST Fashion, (b) Digit MNIST, and (c) Oracle MNIST datasets. The gradual increment in accuracy suggests the efficiency of the network for learning practical features of the image with an optimal convergence pace.
}
    \label{fig:result}
\end{figure*}

\textbf{Mnist Digit Dataset -} The fact that most of the models use this dataset and that it also meets our requirement for a low-resolution image led us to choose it as a way to demonstrate the power of our architecture while providing a good way to compare our results with those of other state-of-the-art models.
The MNIST dataset was created using NIST's Special Databases 3 and 1, which include binary images of handwritten integers. Initially, NIST classified SD-3 as the training set and SD-1 as the test set. SD-3, on the other hand, is much cleaner and easier to discern than SD-1. The MNIST training set contains 30,000 SD-3 patterns and 30,000 SD-1 patterns. Our test set included 5,000 SD-3 patterns and 5,000 SD-1 patterns. The SD-1 has 58,527 digit images authored by 500 different writers. In contrast to SD-3, where blocks of data from each writer appear in sequence, the data in SD-1 is fragmented.

\textbf{Fashion Mnist Dataset -} Fashion-MNIST\cite{xiao2017/online} is a dataset of Zalando article photos, with 60,000 examples in the training set and 10,000 examples in the test set. Each example is a 28x28 grayscale image paired with a label from one of ten categories. Fashion-MNIST is intended to be a drop-in replacement for the original MNIST dataset for evaluating machine learning algorithms. The image size and structure are the same as in the training and testing splits.

Each image is 28 pixels high and 28 pixels wide, for a total of 784 pixels. Each pixel has a single pixel-value associated with it, which indicates how light or dark that pixel is, with larger numbers indicating darker. This pixel value is an integer ranging from 0 to 255. There are 785 columns in the training and test data sets. The class labels are listed in the first column. The first represents the clothing, while the second represents the accessories. The remaining columns contain the corresponding image's pixel values.   

\textbf{Oracle Mnist Dataset - } The Oracle-MNIST dataset \cite{wang2022oracle} contains 30,222 ancient characters from ten categories in $28 \times 28$ grayscale image format for pattern classification, with special challenges in image noise and distortion. The training set has 27,222 images, while the exam set has 300 images per class. It uses the same data structure as the original MNIST dataset, making it compatible with all existing classifiers and systems. However, it is more difficult to train a classification model on it without overfitting and low performance. Images of ancient characters suffer from 1) incredibly serious and unusual noises created by three thousand years of burial and aging, as well as 2) significantly different writing styles in ancient Chinese, both of which make them realistic for machine learning study.

We chose this dataset due to its noisy and low-resolution characteristics. This aspect of the dataset makes it extremely difficult for a standard model to classify the images. Figures x and y show that the Inception v3 \cite{szegedy2015going} and Vgg-16 \cite{simonyan2014very} models are not performing optimally, and we can see an obvious degradation in the models' performance.

\subsection{Training setup}

To improve the reliability of our results, we use the same setup and input size for training and testing all models. The images have been resized to $35 \times 35 \times 1$ in order to preserve the standard input size for models like Inception-v3 \cite{szegedy2015going} and VGG-16 \cite{simonyan2014very}, which don't function properly with $28 \times 28$. For training the models we use Google-Colab with an NVIDIA Tesla T4 GPU with 16GB memory, with a batch size of 256. For analyzing the model's behavior over a long period of time, models were trained for 200 epochs. However, for publishing the model we use a callback function that stops the training after 30 unchanged epochs on validation loss this technique is applied to save time and prevent overfitting.

\subsection{Results}
In this section, the results of our work will be presented and we will have a comparison between our model's accuracy and loss against other models to evaluate the effectiveness of introduced architecture in low-resolution image classification tasks.

% As shown in Fig. \ref{fig:result}, the model has very robust and stable results on the mentioned datasets without any overfitting issues. The model's performance is directly impacted by the image's detail and noise. For example, model accuracy is lower on Oracle-MNIST\cite{wang2022oracle} because the dataset contains a lot of noisy images, and also the inner-variance in the labels affects the performance of the model, but compared to other state-of-the-art models, the introduced method can achieve an acceptable result with an accuracy of 95.13 according to Table \ref{tab:compare_table}. As a result of using a block-based architecture, our model achieved better results. We use residual connections between the input and upper levels of each layer in order to maintain the features. This will enable us to overcome the gradient exploding problem in each block of architecture. One side tries to extract as many global and big features as possible, while the other side tries to find the detailed, local features in images. Using residual connections in combination with different kernel sizes in each layer enables the classification of low-resolution images with high quality.

As shown in Fig. \textcolor{red}{\ref{fig:result}}, our model easily outperformed 3 other state-of-art models in terms of accuracy in all 3 datasets, without experiencing overfitting or a gradient exploding problem. It is important to note that the introduced model is a simpler and easier model than the other two. We can see this by looking at the number of parameters in table \textcolor{red}{\ref{tab:compare_table}}, and despite the lower number of parameters, it outperformed the other famous models due to the unique MK-blocks we introduced in this section \textcolor{red}{\ref{sec:method}}. By looking at the tables and charts one point that we can understand is that the model's performance is directly impacted by the image's detail and noise. For example, the model's accuracy is lower on Oracle-MNIST\cite{wang2022oracle} dataset because the dataset contains a lot of noisy images, and also the inner-variance in the labels affects the performance of the model, but compared to other state-of-the-art models, the introduced method can achieve a higher and acceptable result with an accuracy of 95.13 according to table \textcolor{red}{\ref{tab:compare_table}}. We were expecting these results because of our unique block-based architecture that we introduced in section \textcolor{red}{\ref{sec:method}}. In each block, we used a number of residual connections between the input and upper levels of each layer in order to maintain the features. This turns out to be very effective at increasing the meaningful features of models and also enabled us to overcome the gradient exploding problem that we have in deeper networks. Using residual connections in combination with different kernel sizes in each layer leads to achieving better results, as we can see in both the table \ref{tab:compare_table} and Fig. \textcolor{red}{\ref{fig:result}}.
\begin{figure}[ht]
 ]   \centering
    \includegraphics[width = \columnwidth]{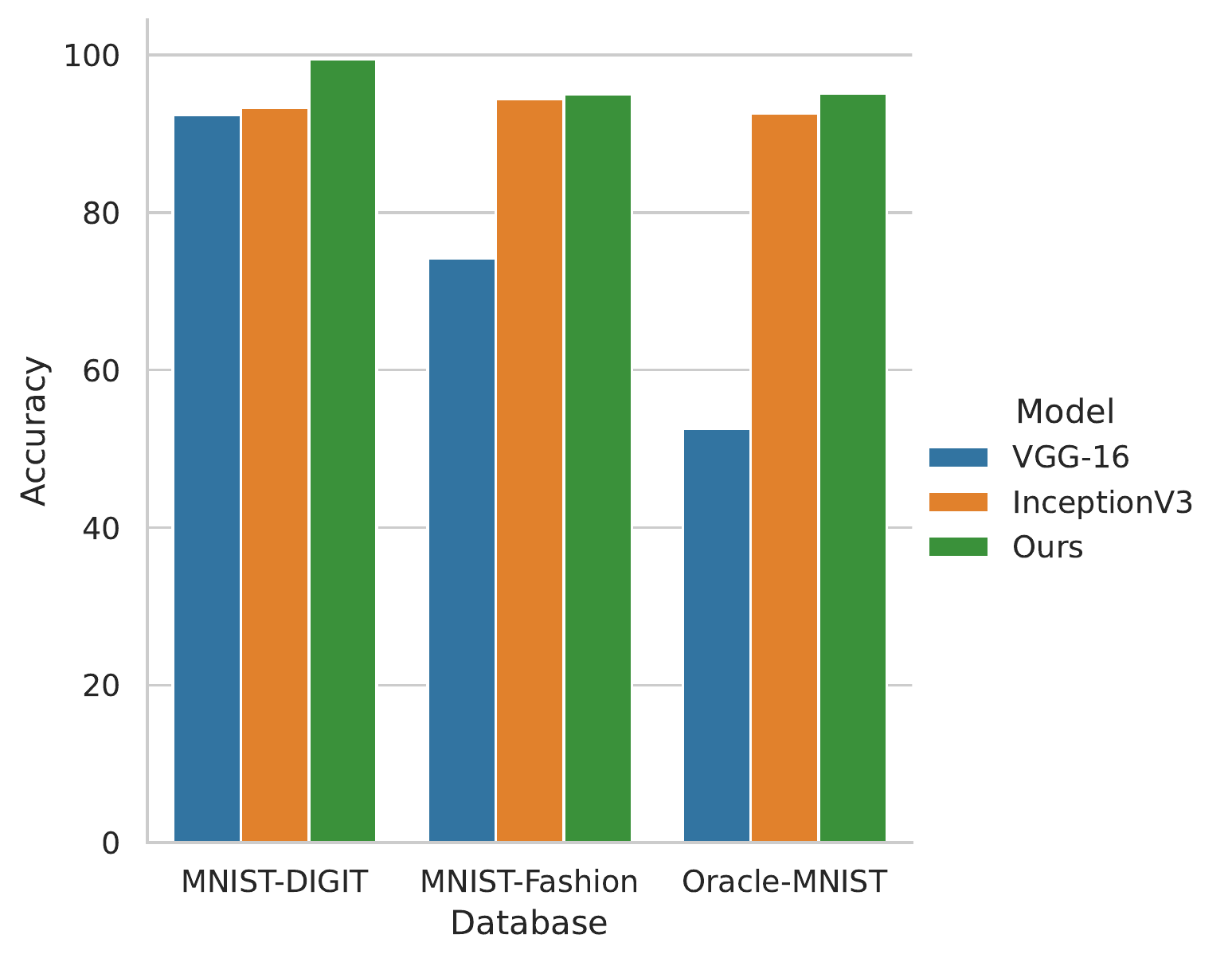}
    \caption{A bar chart illustrating the accuracy of the proposed block-based network. Three datasets are considered: MNIST Digit, MNIST Fashion, and Oracle MNIST.}
    \label{fig:compare_bar}
\end{figure}
\subsection{Comparative Evaluation}

\begin{table}[b]
\centering
\captionsetup{justification=centering}
\caption{The accuracy and number of parameters of 4 different image classification algorithms on 3 famous mnist datasets}
\label{tab:compare_table}
\begin{tabular}{@{}ccccc@{}}
\toprule
\makecell{Model \\Name}    & \makecell{Digit \\MNIST} & \makecell{Fashion \\MNIST\cite{xiao2017/online}} & \makecell{Oracle\\ MNIST\cite{wang2022oracle}} & \makecell{Number of \\Parameters}\\ \midrule
Inception-V3  & 93.31       & 94.44         & 92.6      & 23,851,784 \\
VGG-16        & 92.4        & 74.2          & 52.6      & 138,357,544 \\
Ours          & 99.47       & 95.03         & 95.13     & 1,028,234  \\  \bottomrule

\end{tabular}
\end{table}
The purpose of this section is to compare our methods with VGG-16 \cite{simonyan2014very} and Inception-V3 \cite{szegedy2016rethinking} on three popular datasets: MNIST-Digit, MNIST-Fashion \cite{xiao2017/online}, and Oracle-MNIST \cite{wang2022oracle}. As we discussed earlier in sectioin \textcolor{red}{\ref{sec:method}}, we used the idea of the inception module and the Res-Net connection in order to improve the accuracy and performance of the model on low-resolution images. 

According to Fig. \textcolor{red}{\ref{fig:compare_bar}}, our newly developed model outperforms both other models. We get better results by reducing the number of parameters, which reduces computation overhead and increases speed. As shown in table \textcolor{red}{\ref{tab:compare_table}}, our model is much more accurate than other models in the MNIST dataset with 99.47 accuracy. Because, this dataset is an easy and simple dataset so, we anticipated this kind of performance for this dataset. In the second dataset (fashion-mnist\cite{xiao2017/online}), which is a harder dataset to classify due to the image complexity we have on images, our model can achieve a better accuracy compared to other and by looking at Fig. \textcolor{red}{\ref{fig:result}}, we can see the model keeps its robustness during training.

The real challenge comes with Oracle-MNIST \cite{wang2022oracle}, since it is a newly collected dataset with unique characteristics. As indicated in this section \textcolor{red}{\ref{Datasets}}, there is some natural noise and blur in images. As a result, the images have very poor quality, which will reduce the performance of the models. Based on our tests, which are available in the table. \textcolor{red}{\ref{tab:compare_table}}. The old two state-of-the-art models (inception-v3 \cite{szegedy2016rethinking}, VGG-16 \cite{simonyan2014very}) had very poor performance on Oracle-MNIST \cite{wang2022oracle}, despite their higher number of parameters and complexity of their architecture (table \ref{tab:compare_table}). However, our model achieves a better result on this dataset and it is able to classify images with greater confidence while having less power consumption than inception-v3 \cite{szegedy2016rethinking} and VGG-16 \cite{simonyan2014very} due to its lower number of parameters.

% Fig. \ref{fig:compare_bar} shows that our newly developed model performs better than the two other models. We get better results with a lower number of parameters which will reduce the computation overhead and increase the speed. By looking at table \ref{tab:compare_table} we can view the results of 3 different models on different datasets. Based on table \ref{tab:compare_table}, it is evident that the two state-of-the-art models have very low accuracy on the Oracle-MNIST dataset since this dataset's images are very noisy and some have some natural blur. However, the proposed architecture excels at classification for this dataset over the two state-of-the-art models. According to table \ref{tab:compare_table}, our model has fewer parameters than other models. As a result of their high number of parameters, the inception-v3 and the VGG-16 models use more computation power, and are also more likely to overfit on low-resolution images. The simplicity and speed of our model make it an excellent solution for noisy and low-resolution images.

\section{Conclusion and Future Work}
\label{sec:con}

In this paper, a novel architecture for low-resolution image classification is proposed and examined. Using the results of our study, we can conclude that this model can outperform many of the state-of-the-art models that are currently available in image classification tasks. In addition, the presence of modules similar to those seen at inception may contribute to these results. With the help of these two ideas, we were able to create a model that was simpler and more efficient comparing to others. The model was evaluated on MNIST family datasets and is generalizable to other low-resolution ones.

As a future work, there are several modifications that can be performed to make our model more robust against noises and make it faster by reducing the number of parameters. According to our findings, the image size decreased dramatically after the first MK-block, so having the same number of filters for the second and third MK blocks would not be optimal, which increases computational costs and would be undesirable. The goal is to introduce a new hyperparameter for the number of filters in the second and third blocks in order to make these blocks more flexible for the individual image.
Another way to manage this problem is to use deconvolution to increase the image size after each block. This will increase feature extraction and make the system more robust overall.
\bibliographystyle{ieeetr}
\bibliography{references}

\end{document}